\documentclass{bmvc2k}
\usepackage{amsthm,amsmath,amssymb,bm}
\usepackage{mathrsfs}
\usepackage{multirow}
\usepackage{makecell}
\usepackage{hyperref}

\title{BiHand: Recovering Hand Mesh with Multi-stage Bisected Hourglass Networks}

\addauthor{Lixin Yang}{siriusyang@sjtu.edu.cn}{1}
\addauthor{Jiasen Li}{lijiasen0921@sjtu.edu.cn}{2}
\addauthor{Wenqiang Xu}{vinjohn@sjtu.edu.cn}{3}
\addauthor{Yiqun Diao}{diaoyiqun@sjtu.edu.cn}{4}
\addauthor{Cewu Lu}{lucewu@sjtu.edu.cn}{5}

\addinstitution{
 Machine Vision and Intelligence Group\\
 Shanghai Jiao Tong University\\
 Shanghai, China
}

\runninghead{Student, Prof, Collaborator}{BMVC Author Guidelines}

\def\eg{\emph{e.g}\bmvaOneDot}

\def\etal{\emph{et al}\bmvaOneDot}

\begin{document}

\maketitle

\begin{abstract}


3D hand estimation has been a long-standing research topic in computer vision. A recent trend aims not only to estimate the 3D hand joint locations but also to recover the mesh model. However, achieving those goals from a single RGB image remains challenging. In this paper, we introduce an end-to-end learnable model, BiHand, which consists of three cascaded stages, namely 2D seeding stage, 3D lifting stage, and mesh generation stage. At the output of BiHand, the full hand mesh will be recovered using the joint rotations and shape parameters predicted from the network. Inside each stage, BiHand adopts a novel bisecting design which allows the networks to encapsulate two closely related information (\eg 2D keypoints and silhouette in 2D seeding stage, 3D joints, and depth map in 3D lifting stage, joint rotations and shape parameters in the mesh generation stage) in a single forward pass. As the information represents different geometry or structure details, bisecting the data flow can facilitate optimization and increase robustness. For quantitative evaluation, we conduct experiments on two public benchmarks, namely the Rendered Hand Dataset (RHD) and the Stereo Hand Pose Tracking Benchmark (STB). Extensive experiments show that our model can achieve superior accuracy in comparison with state-of-the-art methods, and can produce appealing 3D hand meshes in several severe conditions. The training codes, model and dataset are publicly available at \url{https://github.com/lixiny/bihand}.

\end{abstract}

\section{Introduction}
\label{sec:intro}

Hand is one of the most crucial elements when human interact with surroundings. Extracting 3D structure of a hand from a single RGB image can benefit plenty of applications including VR/AR, action recognition, and human-computer interaction. Generally, 3D hand estimation is formulated as 3D joint location estimation, or more detailed 3D mesh reconstruction. Since 3D hand mesh contains richer geometric information and is considered indispensable for applications involving grasping or hand-object contact, in this work, we focus on recovering the 3D hand model from a single RGB image.

The past decade has witnessed rapid growth in the study of 3D hand mesh recovery. Early works applied multi-view stereo approaches to reconstruct hand \cite{ueda2001hand, guan2006multi}. Later, to reduce the requirement of multi-view images, model templates (\eg Primitives approximation \cite{oikonomidis2011efficient}, sphere-meshes \cite{tkach2016sphere}, MANO \cite{romero2017embodied}) were adopted to fit the prior hand shape to a single image. With the advent of deep learning techniques, it became possible to directly estimate hand model parameters from color or depth pixels \cite{malik2018deephps, panteleris2018using, boukhayma20193d}. Recently, RGB-based learning methods became the mainstream approach. The majority of those methods \cite{boukhayma20193d, zhang2019end, hasson2019learning, baek2019pushing} learned to regress the parameters of hand model, typically MANO.  However, since they regressed the lossy compressed PCA component of MANO, their accuracy cannot be guaranteed. To address this, an inverse kinematics based method \cite{zhou2020monocular} was proposed to calculate the rotation angle from the location by leveraging the kinematic chains of hand.

\begin{figure}
\begin{center}
\vspace{3 pt}
\includegraphics[width=\linewidth]{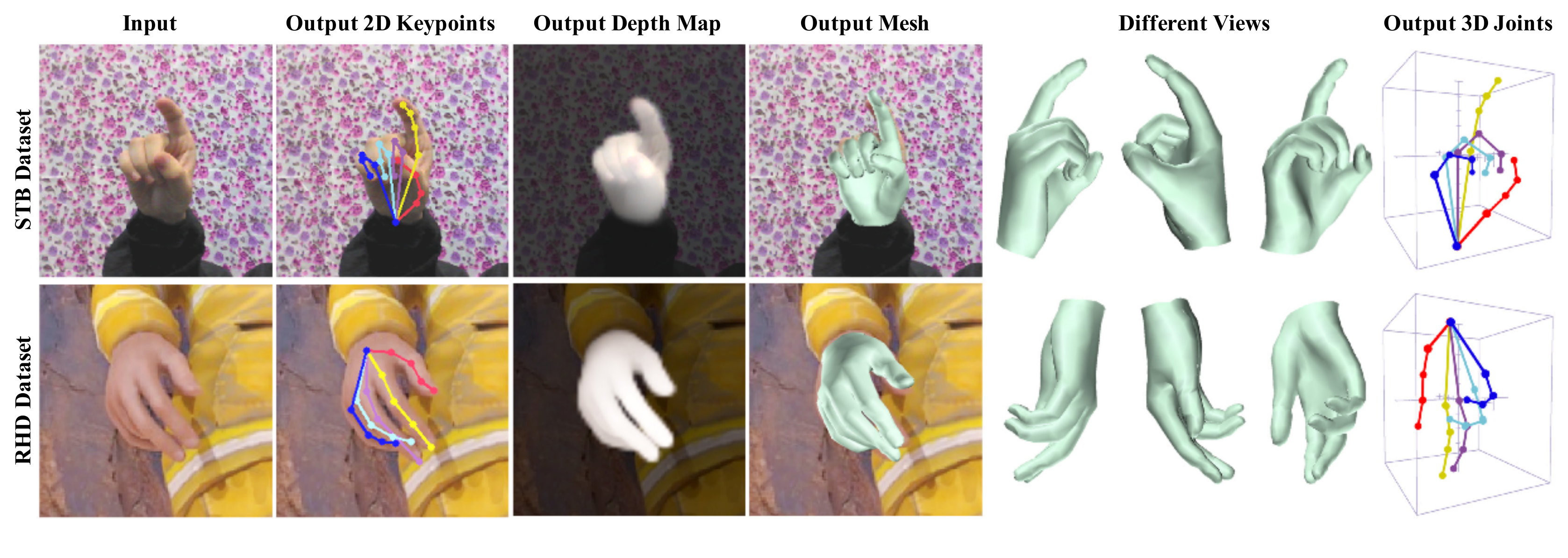}
\end{center}
\vspace{-10 pt}
\caption{We introduce a multi-stage bisected network, BiHand, for single RGB image hand mesh recovery. BiHand progressively converts RGB image to 2D geometry (2D keypoints and silhouette), then to 3D structure (3D joints and depth map), and finally to a full hand mesh in a single forward pass.  }
\vspace{-8 pt}
\label{fig:teaser}
\end{figure}

Estimating a 3D model from a 2D image is intrinsically an ill-posed problem. Specifically, it suffers from the perspective ambiguity. To alleviate this issue, besides adopting CNN as the main feature extractor to estimate the hand model parameters, former researches also incorporated more geometric information as auxiliary supervision, such as 2D keypoints, silhouette, 3D joints, depth map, etc. Such additional geometric information can be used as intermediate regularization
or post regularization.
However, how to arrange these information is non-trivial. 2D keypoints and silhouette both represent planar projection with different level of details. They are strong cues to infer the 3D structure. Similarly, 3D joints and depth map encode 3D structure information with different levels of details. When it comes to estimate hand model, rotation parameters and shape parameters also belong to different information flow. Based on this observation, We propose a multi-stage framework with bisecting design, named \textbf{BiHand}. As is shown in Fig. \ref{fig:teaser}, it can progressively convert the RGB image to 2D geometry within the 2D seeding module (SeedNet), then to 3D structure within the 3D lifting module (LiftNet), and finally to the mesh model within mesh generation module with a novel shape-aware inverse kinematics network (SIKNet).

To train SIKNet requires joint location-rotation training pair.
However, only a few MoCap samples are available in the MANO dataset. Thus, we create a SIK-1M dataset containing one million synthetic location-rotation pairs. Though the synthesis can potentially generate unlimited pairs of training data, we find one million is enough.

For quantitative evaluation, we experiment our method on two standard benchmarks, namely the Rendered Hand Dataset (RHD)\cite{zimmermann2017learning} and the Stereo Hand Pose Tracking Benchmark (STB) \cite{zhang2017hand}. BiHand achieves state-of-the-art performance with AUC 0.951 on the RHD and 0.997 on the STB.

In summary, our contributions are as follows:
\begin{itemize}
   \item {We propose a new end-to-end learnable framework, BiHand, to address the single RGB image hand mesh recovery problem. BiHand leverages both planar and 3D structural information as intermediate representations to stabilize the training. A novel bisecting design is proposed to organize the geometric information flow.}
   \item {To recover the mesh from estimated 3D joints, we also propose a shape-aware inverse kinematics network, SIKNet, to map the joint locations to MANO parameters. To train SIKNet with full supervision, we build up a large-scale dataset, SIK-1M.}
\end{itemize}

\section{Related Work}

Our method closely relates to 3D hand pose estimation and 3D hand mesh reconstruction problems.

\subsection{3D Hand Pose Estimation}
Early works on 3D hand pose estimation mainly focused on regressing hand joints from a single depth image. They either exploited a model fitting \cite{Khamis_2015_CVPR} or learned a depth-joint mapping \cite{Sun_2015_CVPR}. Recently, several deep learning based methods improved the estimation by employing CNNs \cite{tompson2014real}, multi-view CNN \cite{ge2016robust}, 3D CNN \cite{ge20173d} or PointNet \cite{qi2017pointnet, ge2018hand}. Another stream on this topic is to estimate the 3D hand pose from an RGB image. Former researches adopted CNN feature extractor with several learning strategies including variational auto-encoder \cite{Spurr_2018_CVPR}, iterative skeleton fitting \cite{mueller2018ganerated}, depth regularizer \cite{cai2018weakly, chen2020dggan}, etc.

\subsection{3D Hand Mesh Reconstruction}

\paragraph{Multi-view Stereo Based Methods}
Since early single-view algorithms were sensitive to occlusion and noise, several multi-view methods were put forward. Delamarre and Faugeras \cite{delamarre1998finding} proposed to fit finger models on the stereo-based scene reconstruction. Ueda \etal \cite{ueda2001hand} exploited multi-view silhouette images to reconstruct hand as a voxel model.  Guan \etal \cite{guan2006multi} formulated this problem by fusing the multi-view images into a maximum a posterior framework. However, these methods usually require multi-camera setups, which is considered less general than a single depth or RGB camera nowadays.

\paragraph{Depth Based Methods}
Recovering meshes from a single depth image was mostly regarded as a model fitting problem. Khamis \etal \cite{Khamis_2015_CVPR} proposed to fit a morphable hand model into depth image by linear blender skinning (LBS). LBS function requires two sets of parameters, i.e. poses and shapes, to articulate a hand model. Remelli \etal \cite{Remelli_2017_ICCV} proposed to decouple pose and shape parameters and calibrate the shapes in a low dimensional space. A recent method \cite{malik2018deephps} proposed to directly regress LBS parameters using CNN on the depth image.

\paragraph{RGB Based Methods}
Previous methods focused on fitting a rigid articulated hand model on an RGB image. Paschalis \etal \cite{panteleris2018using} proposed to estimate hand joint position and adopted iterative model fitting to obtain the joint rotations. Similar ideas were also proposed by Kokic \etal \cite{kokic2019learning}. However, those methods are incapable of capturing hand shape variants. Recently, with a parametric hand model MANO, several researchers proposed to fit its pose and shape parameters into pixels. Boukhayma \etal \cite{boukhayma20193d} first proposed to directly regress the MANO parameters from the input of image and heatmaps. Zhang \etal \cite{zhang2019end} forwarded it by adopting a neural render to employ silhouette supervision on meshes. Adopting neural render to supervise mesh was also proposed in \cite{ge20193d, baek2019pushing}. Since both Boukhayma and Zhang's methods only regressed the MANO's PCA components, their accuracy suffered from its lossy property. Recently, instead of regressing on PCA, Zhou \etal \cite{zhou2020monocular}  proposed to directly estimate the rotations of all hand joints. Zhou's methods can achieve higher robustness on both joint and mesh estimation. Instead of using MANO, Ge \etal \cite{ge20193d} proposed to directly regress the vertex positions on a template hand mesh using GraphCNN \cite{defferrard2016convolutional}. However, their method requires training on a dataset with dense vertices annotations, which is not practical in the real world.
Compared with most of the aforementioned methods, our BiHand is more informative and feasible. The bisected hourglass allows the networks to encapsulate homogeneous information during training, and the decoupled stages make the training process more effective.

\begin{figure*}
   \begin{center}
   \vspace{5 pt}
   \includegraphics[width=1.0\linewidth]{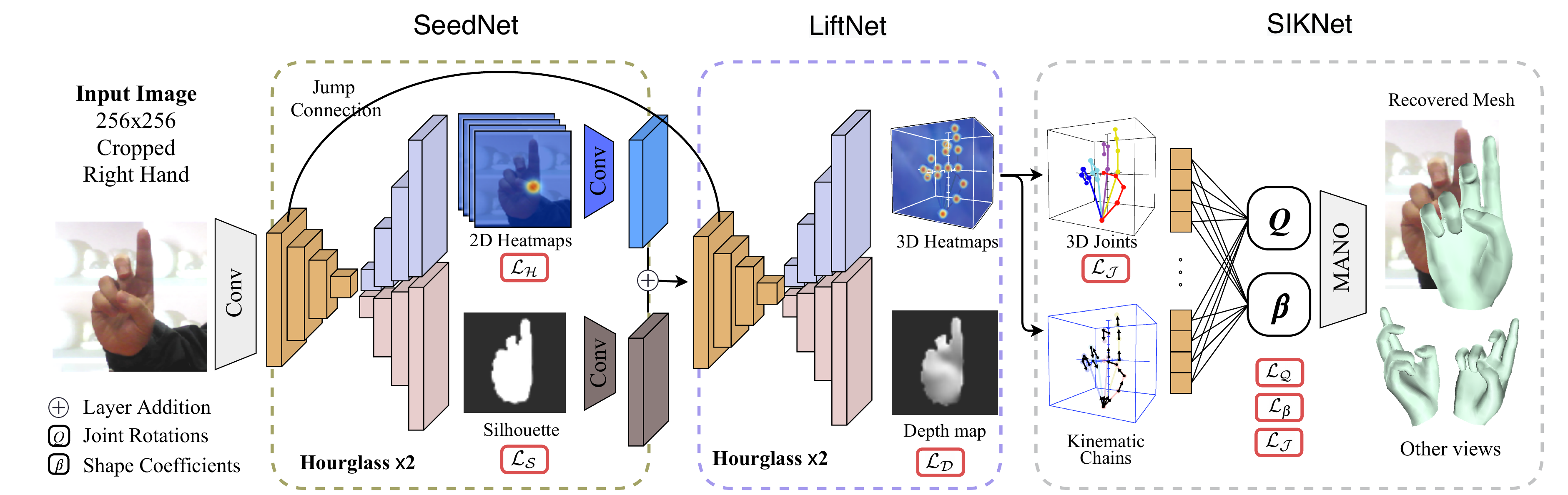}
   \end{center}
      \vspace{-5 pt}\caption{ Overview of our proposed multi-stage bisected network (BiHand) on 3D hand mesh reconstruction. It comprises three sequential modules. First, the 2D seeding module, SeedNet,  predicts 2D heatmaps and silhouette. Second, the 3D lift module, LiftNet, lifts the 2D predictions to 3D heatmaps and depth map. Third, the mesh generation module with SIKNet recovers the full hand meshes based on the 3D predictions. }
      \vspace{-8 pt}
   \label{fig:architecture}
\end{figure*}

\vspace{-5 pt}\section{Method}
\subsection{Overview}
The proposed BiHand seeks to predict a 3D hand mesh model from a single RGB image.
To achieve this, given an image $\mathcal{I}$ as input, BiHand first predicts 2D heatmaps $\mathcal{H}_{2D} \in \mathbb{R}^{K \times H \times W}$ and silhouette $\mathcal{S} \in \mathbb{R}^{H \times W}$ through 2D seeding module, SeedNet (Sec. \ref{sec:2dtectnet}). Then  2D geometries are lifted to 3D heatmaps $\mathcal{H}_{3D} \in \mathbb{R}^{K \times Z \times H \times W}$ and depth map $\mathcal{D} \in \mathbb{R} ^ {H \times W}$ through 3D lifting module, LiftNet (Sec. \ref{sec:3dtectnet}). And thus 3D joint locations $\mathcal{X}\in \mathbb{R}^{K \times 3}$ can be obtained from $\mathcal{H}_{3D}$, and be put through the mesh generation module with SIKNet (Sec. \ref{sec:SIKNet}) to recover the full hand mesh $\mathcal{M}(\theta, \beta) \in \mathbb{R}^{N \times 3}$, where $\theta \in \mathbb{R}^{16 \times 3}$, $\beta \in \mathbb{R}^{10} $ are the rotation and shape parameters from MANO.
The overall pipeline is illustrated in Fig. \ref{fig:architecture}.

If not specified, $K =21$ indicates the number of hand joints and $N=778$ indicates the number of mesh vertices. $(H, W) = (64, 64)$ means the resolution of 2D heatmaps, silhouette, and depth map. $Z=64$ is the depth resolution of 3D heatmaps.

\begin{figure*}
   \begin{center}
   \includegraphics[width=0.7\linewidth]{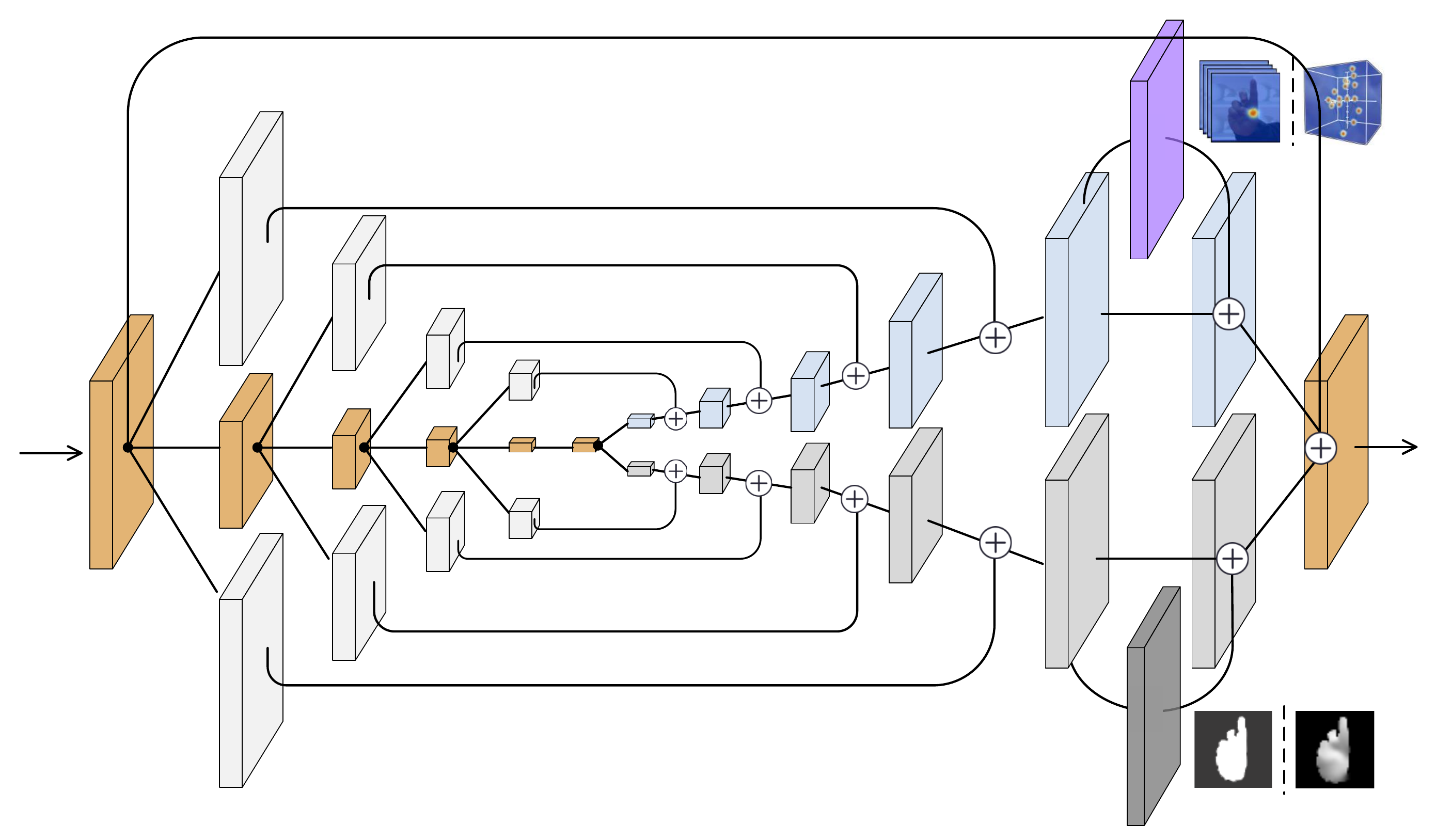}
   \end{center}
      \vspace{-15 pt}\caption{ Illustration of a single bisected hourglass network. Each block represents a residual module. The resolutions of input and output layer are both 64 $\times$ 64 and the number of channels across the whole network is 256.  }
   \label{fig:2bhourglass}
\vspace{-10 pt}
\end{figure*}

\vspace{-5 pt}
\subsection{2D Seeding Module, SeedNet}
\label{sec:2dtectnet}
The first stage of BiHand is 2D seeding module, which transforms the RGB image $\mathcal{I}$ to 2D heatmaps $\mathcal{H}_{2D}$ and 2D silhouette $\mathcal{S}$. Adopting the 2D information as the bridge between RGB and 3D structure has three advantages: interpretable,  robust, and easy to train. Some studies on 3D hand pose estimation confirmed the observation \cite{mueller2018ganerated,mueller2017real,zimmermann2017learning,cai2018weakly,chen2020dggan}.

$\mathcal{H}_{2D}$ and $\mathcal{S}$ both represent the 2D structure with different levels of details, which is double-edged. For the up-side, they can be encoded simultaneously since the source is the same, and when $\mathcal{H}_{2D}$ is sensitive to motion blur or self-occlusion, $\mathcal{S}$ can provide stable cues. For the down-side, they should be separately treated when decoding, since the amount of information to be restored are different. Of course, we can leverage the high capacity of CNN to force a single decoder to encapsulate two different information flows, but it is easier to just bisect the flow with two decoders to learn.

\vspace{-8pt}
\paragraph{Bisected Hourglass Network}
To separate the information flow, SeedNet is designed as a stacked bisected hourglass network, as shown in Fig. \ref{fig:2bhourglass}. Bisected hourglass means the network has one encoder, two decoders where the encoder and decoder are the same with hourglass \cite{newell2016stacked}. Specifically, it first encodes the input to latent features $\mathcal{F}_{2D}$ through 4 downsampling blocks.
The latent features $\mathcal{F}_{2D}$ are then passed through two separated decoders, one for heatmaps $\mathcal{H}_{2D}$ and the other for silhouette $\mathcal{S}$. The decoder also consists of 4 upsampling blocks. Each pixel in the $k^{\rm th}$ heatmap $\mathcal{H}_{2D}^{(k)}$ indicates the confidence of that pixel being covered by the  $k^{\rm th}$ keypoint.
And each pixel in silhouette $\mathcal{S}$ indicates the confidence of that pixel belongs to hand's segmentation.
The two decoders are the same in architecture and symmetrical to the encoder, but they do not share the same weight. Similar to the original hourglass, the bisected hourglass can be stacked as well. We stack two bisected hourglasses in SeedNet. For each hourglass, we add intermediate supervision on the estimated 2D heatmaps and silhouette, which are then concatenated as the input of the subsequent networks.

\vspace{-8pt}
\paragraph{Loss Terms}
The loss function of SeedNet
\begin{equation}
   \mathcal{L}_{2D} = \lambda_{\mathcal{H}} \ \mathcal{L}_{\mathcal{H}} + \lambda_{\mathcal{S}} \ \mathcal{L}_{\mathcal{S}}
   \label{lossSeedNet}
\end{equation}
comprises two terms. First, the heatmaps loss $\mathcal{L}_{\mathcal{H}}$ is defined as the pixel-wise mean squared error (MSE) between the estimated and ground-truth 2D heatmaps. The ground-truth 2D heatmap $\mathcal{H}_{2D}^{(k)}$ of keypoint $k$ is generated by 2D Gaussian distributions with mean at $k$'s annotations and standard deviation $\sigma=1.5$. The second term silhouette loss $ \mathcal{L}_{\mathcal{S}} $ is the cross-entropy loss between the predicted and the ground-truth  silhouette.

\vspace{-5 pt}
\subsection{3D Lifting Module, LiftNet}
\label{sec:3dtectnet}
The objective of LiftNet is to infer the 3D joint location given the 2D heatmaps and silhouette from SeedNet. Several solutions \cite{zimmermann2017learning, tome2017lifting} were proposed to directly lift the estimated 2D heatmaps to 3D coordinates. However, as was argued in \cite{cai2018weakly}, one can hardly resolve the inherent ambiguity in projection. Similar to \cite{cai2018weakly}, we alleviate this issue by maintaining the contextual information alongside with heatmaps in the network. Besides, We choose to estimate 3D heatmaps instead of joint vectors ($x,y,z$ coordinates of $K$ joints) since the regression on discrete coordinates imposes high non-linearity to this problem.

Similar to 2D information, 3D heatmaps and depth map are homogeneous in terms of 3D spatial structures.
Training the depth map predictor could provide sufficient information on the relative depth to prevent joint predictors from abnormalities.  However, 3D heatmaps and depth map are also different in the grained level. Therefore, we also adopt the bisected design in LiftNet.

\vspace{-8pt}
\paragraph{LiftNet Design}
The architecture of our LiftNet follows SeedNet, while changing the output from 2D heatmaps, silhouette to 3D heatmaps, depth map.
Our joint decoder tackles 3D joint estimation as classification by outputting $K$ volumetric blocks (3D heatmaps)
$\mathcal{H}_{3D}$. Each voxel in $k^{\rm th}$ block $\mathcal{H}_{3D}^{(k)}$ represents the likelihood of this voxel being covered by the $k^{\rm th}$ joints.
We design the joint decoder to output 3D heatmaps in the normalized-$uvd$ form, where $uv$ corresponds to UV coordinates on the input image and $d$ to the joint's root-relative scale-invariant depth value. The volumetric 3D heatmaps can be transferred to $uvd$ coordinates through a differentiable $argmax$ function. Technically,  for $j^{\rm th}$ joints, its $uvd$ coordinates can be obtained by:
\vspace{-5 pt}
\begin{equation}
   \vspace{-5 pt}
   [u_j, \ v_j, \  d_j]^\top = \sum_{d=1}^{Z} \sum_{v=1}^{H} \sum_{u=1}^{W} \mathcal{H}_{3D}^{(j)}(u,v,d) \cdot [u, \ v, \ d]^\top
    \label{soft-argmax}
\end{equation}
where $\sum_{u,v,d} \mathcal{H}_{3D}^{(j)}(u,v,d) \equiv 1$.  Finally, the  $uvd$ coordinates are converted to joint locations $\mathcal{X}^{(j)} = (x_j, y_j, z_j)^{\top}$ using camera intrinsic matrix.  Alongside with the joint decoder, the depth decoder predicts the normalized depth map $ \mathcal{D}$,  where each pixel in $\mathcal{D}$ represents the normalized depth value of that pixel. Similar to SeedNet, we stack two bisected hourglass in LiftNet. Inspired from \cite{pavlakos2017coarse}, we adopt a coarse-to-fine 3D heatmaps predictor with the resolution along $z$-axis $Z=32$ in the first hourglass and $Z=64$ in the second. Exploiting depth during 3D joint estimation was previously proposed in \cite{cai2018weakly, chen2020dggan, ge20193d}. But 
in their methods the depth predictor was used as a regularizer after joint estimation, while in ours as a parallel counterpart.

\vspace{-10pt}
\paragraph{Loss Terms}
The loss function of LiftNet
\begin{equation}
   \mathcal{L}_{3D} = \lambda_{\mathcal{J}} \ \mathcal{L}_{\mathcal{J}} + \lambda_{\mathcal{D}} \ \mathcal{L}_{\mathcal{D}}
   \label{lossLiftNet}
\end{equation}
consists of two terms. First, the joint loss $\mathcal{L}_{\mathcal{J}}$  is defined as MSE loss on $K$ joint positions: $\mathcal{L}_{\mathcal{J}} = \sum_{j=1}^{K} \| \mathcal{X}^{(j)} - \mathcal{X^*}^{(j)} \|^{2}_2$, where $\mathcal{X^*}$ is the ground-truth joint location. Second, the depth map loss $\mathcal{L}_\mathcal{D}$ is defined as the smooth L1 loss between the estimated and ground-truth depth map: $\mathcal{L}_\mathcal{D} = |\mathcal{D} - \mathcal{D^*}|_{smoothL1}$, where $\mathcal{D^*}$ is the normalized ground-truth depth map.

\vspace{-5 pt}
\subsection{Mesh Generation Module with SIKNet}
\label{sec:SIKNet}
In this section, we recover the full hand mesh by employing a parametric hand model, MANO \cite{romero2017embodied}. MANO is a statistical model based on the SMPL for the human body \cite{loper2015smpl}. Similar to SMPL, MANO formulates an articulated hand $\mathcal{M}$ with pose parameters $\theta$ and shape parameters $\beta$. Technically, $\theta$ represents the 16 joint rotations, excluding 5 fingertips, in the form of axis-angle vectors, and $\beta$ represents the shape PCA coefficients learned from various hand scans.
Similar to the majority of CG characters, MANO drives the articulated hand through the joint rotations.

Previous MANO-based methods in \cite{zhang2019end, boukhayma20193d, hasson2019learning} directly regressed the PCA component. Since PCA is a lossy-compression method, in theory, joint information cannot be perfectly recovered. On the other hand, the human hand is by nature a kinematic chain. Therefore we can convert joint locations to its rotations through inverse kinematics (IK). Zhou \etal \cite{zhou2020monocular} designed an IK network to estimate $\theta$ and applied iterative optimization to obtain $\beta$. By observing that $\theta$ and $\beta$ is closely related, we modify the original IK network to a bisected design. Since the network can simultaneously produce joint rotations and shape coefficients, we name it as shape-aware IK network (SIKNet).

\vspace{-5pt}
\paragraph{Parametric Hand Model}
With the joint rotations and shape coefficients obtained, We can utilize MANO to reconstruct the full hand meshes.
MANO first deforms a mean template mesh $\bar{\mathbf{T}}$ through two blend functions $\mathcal{B}(\theta)$ and $\mathcal{B}(\beta)$, and then the deformed template mesh $\mathcal{T}(\theta, \beta)$ is transferred to final mesh $\mathcal{M}(\theta, \beta)$ through a linear blend skinning (LBS) function $W(\cdot)$:
\begin{equation}
  \mathcal{M}(\theta, \beta) = W(\mathcal{T}(\theta, \beta), \theta, \mathcal{W}, \mathcal{J}(\beta))
  \label{manolayer}
\end{equation}
where $\mathcal{W}$ is blend weights, and $\mathcal{J}(\beta)$ is joint locations. Following the method in \cite{hasson2019learning}, we integrate MANO as a differentiable layer in our SIKNet.

\vspace{-5pt}
\paragraph{SIKNet Design}
SIKNet mainly consists of a bisected regression network: one for rotation regression ($\theta$-Reg) and the other for shape regression ($\beta$-Reg). Both $\theta$-Reg and $\beta$-Reg are defined as network with seven fully-connected layers. We construct the input of $\theta$-Reg as $\mathcal{I} = [\bar{\mathcal{X}}, \bar{\mathcal{K}}] \in \mathbb{R}^{2 \times 21 \times 3}$, where $\bar{\mathcal{X}}$ stands for the root-relative, scale-invariant joint locations and $\bar{\mathcal{K}}$ is the normalized direction vectors along the hand's kinematic chain. Based on the discussion in \cite{zhou2020monocular}, we design $\theta$-Reg to output quaternion ${\mathcal{Q} \in \mathbb{R}^{16 \times 4}}$ instead of axis-angle.
Another bisection in SIKNet is the $\beta$-Reg, which also takes as input $\mathcal{I} = [\bar{\mathcal{X}}, \bar{\mathcal{K}}]$ and outputs the shape parameters $\beta$.

\begin{figure*}
   \begin{center}
   \includegraphics[width=0.85\linewidth]{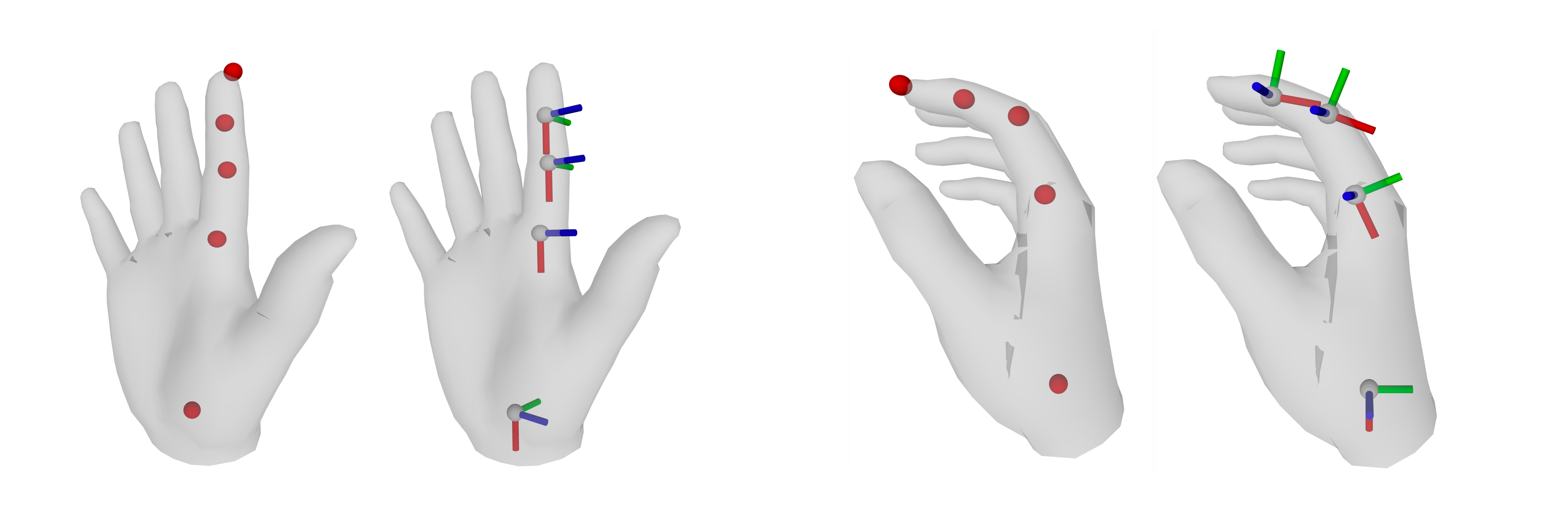}
   \end{center}
   \vspace{-12 pt}\caption{The Inverse Kinematics Dataset. We show two samples of the location-rotation data pair along the index finger. For each sample, the left column shows the 3D joint locations and the right column illustrate the relative joint rotations.}
   \vspace{-10 pt}\label{fig:ikdataset}
\end{figure*}

\vspace{-5pt}
\paragraph{Inverse Kinematics Dataset}
Noticing that the training of $\theta$-Reg and $\beta$-Reg is decoupled from LiftNet, we can, therefore, train the whole SIKNet on a large hand-crafted dataset with direct quaternion supervision and indirect joint supervision. We construct our synthetic dataset through randomly sampling 20k pose and shape parameters with Gaussian distributions $\mathcal{N}(\mu_F, (2 \sigma_F)^2)$, where $\mu_F$ and $\sigma_F$ denote the mean and standard deviation of hands in FreiHand dataset \cite{zimmermann2019freihand} respectively. For each hand ($\theta_i$, $\beta_i$), we then randomly sample 50 camera views. Thus the whole synthetic dataset contains one million hands with ground-truth joint location-rotation data pair (See Fig. \ref{fig:ikdataset}). We split $80\%$ of the synthetic SIK dataset for training and the remaining $20\%$ for testing.

\vspace{-5pt}
\paragraph{Loss Terms}
The loss function of SIKNet
\begin{equation}
   \mathcal{L}_{IK} = \lambda_{\mathcal{Q}} \ \mathcal{L}_{\mathcal{Q}} + \lambda_{\mathcal{\beta}} \ \mathcal{L}_{\mathcal{\beta}} + \lambda_{\mathcal{J}} \ \mathcal{L}_{\mathcal{J}}
   \label{loss2SIKNet}
\end{equation}
consists of three terms. First, same as in \cite{zhou2020monocular}, the quaternion loss $\mathcal{L}_{\mathcal{Q}}$ is a combination of another three terms: cosine loss, L2 loss, and norm loss. The cosine and L2 loss measures the angular cosine error and Euclidean error between the predicted and ground-truth quaternions, respectively, and the norm loss performs regularization.

The second term in Eq. \ref{loss2SIKNet} is the shape loss, which is defined as:
\vspace{3pt}
\begin{equation}
   \mathcal{L}_{\mathcal{\beta}} = \sum_{b} \|\bar{L_b}(\mathcal{Q}, \beta) - \bar{L_b^*} \|^2_2 + \| \beta \|^2_2
   \label{lossshape}
   \vspace{-5pt}
\end{equation}
where $ \bar{L_b}(\mathcal{Q}, \beta)$ and $\bar{L_b^*}$ in first term are the estimated and ground-truth scale-invariant bone length for bone $b$ respectively, and the second term is L2 regularization.

The third term in Eq. \ref{loss2SIKNet} is the joint MSE loss between the estimated joint positions $X(\mathcal{M}(\mathcal{Q}, \beta))$ and ground-truth $\mathcal{X^*}$, where $\mathcal{M}(\cdot)$ is the mesh and $X(\cdot)$ stands for interpolation from meshes to joints. With $\mathcal{L}_{\mathcal{J}}$, we can indirectly supervise the $\theta$-Reg and $\beta$-Reg when training on datasets without ground-truth quaternions annotation.


\vspace{-5 pt}\section{Experiment}


\subsection{Datasets and Metrics}

\label{sec:dataset}
\paragraph{Datasets}
We mainly train and evaluate our network on two public datasets: the Rendered Hand Dataset (RHD) \cite{zimmermann2017learning} and the Stereo Hand Pose Tracking Benchmark (STB) \cite{zhang2017hand}. The RHD is a synthetic dataset with 41258 training and 2728 testing samples. The STB is a real-world dataset containing 12 sequences with 1500 frames per-sequence. We choose 10 sequences in STB for training and the remaining 2 for testing the same as in \cite{zimmermann2017learning}. Similar to \cite{cai2018weakly}, we shift the root joints in STB from palm to wrist to make it consistent with RHD. In both datasets, we mirror all left hands to the right.


\paragraph{Metrics}
We report the percentage of correct 3D keypoints (PCK) and the area under the PCK curve (AUC) as two main evaluation metrics.
The distance thresholds of PCK ranges from 20 $mm$ to 50 $mm$.

\subsection{Implementation Details}
\label{sec:imp}
We implement our framework using PyTorch and train it on two NVIDIA GTX 1080Ti graphic cards.
The network's weights are initialized by \textit{kaiming\_normal} in PyTorch.
For both datasets, we crop the image to be centered at hand and resize it to the input resolution
$256\times256$. During training, input images from RHD and STB are mixed into the same batch.

The training process falls under the multi-task learning scheme.
The hyper-parameters $\{\lambda_{\mathcal{H}}, \lambda_{\mathcal{S}}, \lambda_{\mathcal{J}},
\lambda_{\mathcal{D}}, \lambda_{\mathcal{Q}}, \lambda_{\beta} \} $ are empirically set as
$ \{100, 1, 1000, 1, 1, 1 \}$ to balance different types of supervision.
In our experiment, we first train SeedNet for 100 epochs,
and then exploit its outputs to train LiftNet for another 100 epochs.
In the meantime, we train SIKNet on SIK-1M dataset for 100 epochs. We use Adam optimizer and start with an initial learning rate of $10^{-4}$ among all experiments. The learning rate decreases to $10^{-5}$ at epoch 50.
Finally, we fine-tune the SIKNet on the predicted 3D joints from the LiftNet.
Since no ground-truth joint rotation is provided in STB and RHD dataset, only the shape loss $\mathcal{L}_{\beta}$ and joint loss $\mathcal{L}_{\mathcal{J}}$ are used during fine-tuning. The fine-tuning process starts with the leaning rate of $10^{-5}$ and lasts for 50 additional epochs.

\begin{figure*}
   \vspace{5pt}
   \begin{center}
   \includegraphics[width=1.0\linewidth]{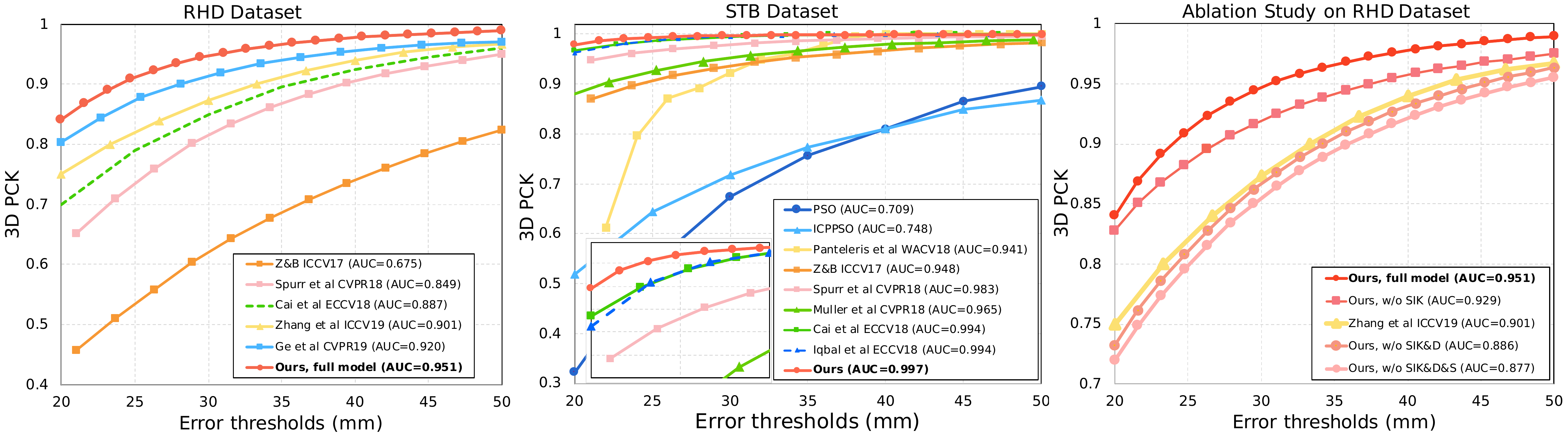}
   \end{center}
   \vspace{-5pt}
   \caption{Quantitative results of BiHand in terms of PCK. The left and middle columns show the comparisons with state-of-the-art methods on RHD and STB dataset. The right column shows the ablation studies on RHD dataset. }
   \label{fig:quantitative}
\end{figure*}

\begin{table}\footnotesize
   \vspace{5pt}
   \begin{center}
      \begin{tabular}{|c|c|c|c|c|c|c|c|c|c|c|}
         \hline
         \multicolumn{2}{|c|}{Method}
            & {\bf Ours}
            & \makecell{Zhou \\ \cite{zhou2020monocular}}
            & \makecell{Ge \\  \cite{ge20193d}}
            & \makecell{Zhang \\ \cite{zhang2019end}}
            & \makecell{Yang \\ \cite{yang2019aligning}}
            & \makecell{Baek \\ \cite{baek2019pushing}}
            & \makecell{Boukhayma \\  \cite{boukhayma20193d}}
            & \makecell{Xiang \\ \cite{xiang2019monocular}}
            \\ \hline
         \multirow{2}{*}{AUC}  & STB  & .997  & .991 & {\bf .998} & .995  & .996 & .995 & .994 & .994   \\ \cline{2-10}
                               & RHD  & {\bf .951}  & .893 & .920 & .901  & .943 & .926 & -   & -   \\ \hline
      \end{tabular}
   \end{center}
   \caption{Additional comparisons with state-of-the-art methods in terms of AUC. " - " denotes methods that did not report the results. }
   \label{table:auc}
\end{table}

\subsection{Results}
\begin{figure*}
   \vspace{3pt}
   \begin{center}
   \includegraphics[width=1.0\linewidth]{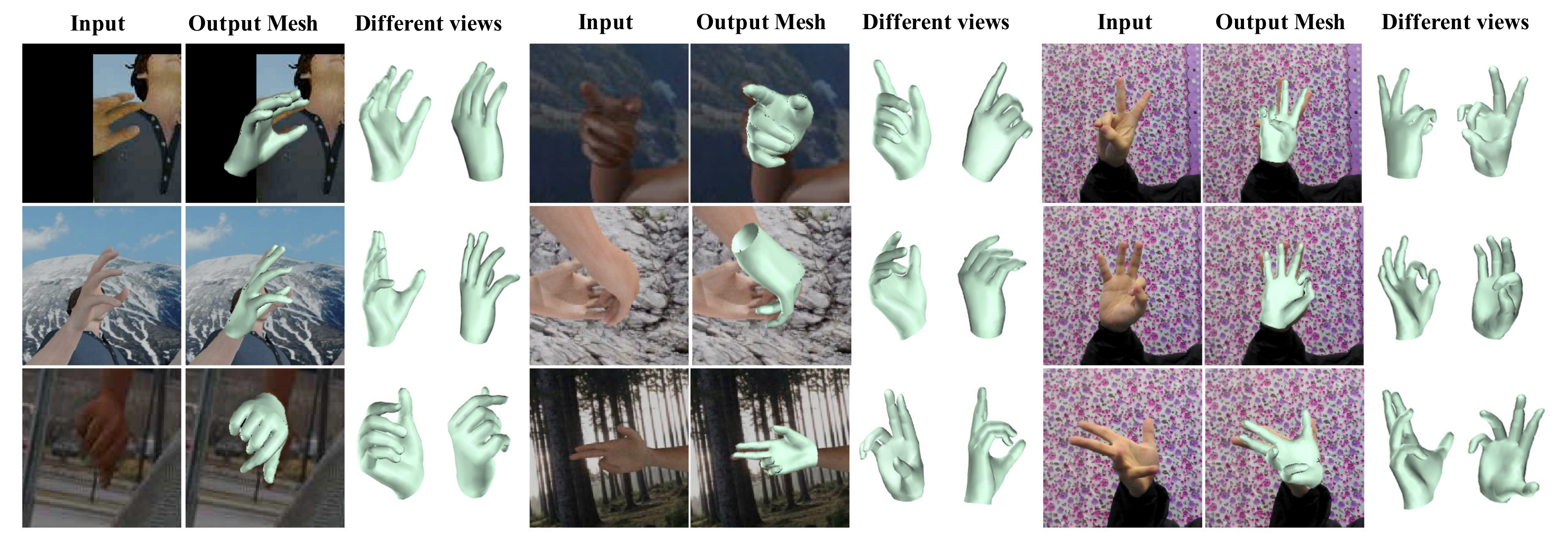}
   \end{center}
   \vspace{-10 pt}\caption{Recovered meshes on RHD (left, middle) and STB (right) dataset.}
   \label{fig:qualitative}
\vspace{-5 pt}\end{figure*}

We evaluate our method both quantitatively and qualitatively.
For quantitative results, we compare our methods in terms of PCK and AUC with
state-of-the-art methods on both RHD and STB dataset.
As is shown in Fig. \ref{fig:quantitative} and and Table \ref{table:auc},
on RHD dataset, our method outperforms those methods in
\cite{zhou2020monocular,ge20193d,zhang2019end,yang2019aligning,baek2019pushing,cai2018weakly,zimmermann2017learning}
over all error thresholds.
On STB dataset, our method is also competitive with those in \cite{zhou2020monocular,ge20193d,zhang2019end,yang2019aligning,baek2019pushing,cai2018weakly,zimmermann2017learning, xiang2019monocular, boukhayma20193d, iqbal2018hand, Spurr_2018_CVPR, mueller2018ganerated, panteleris2018using}.
We also show the qualitative results through the generated mesh in Fig. \ref{fig:qualitative}.
We intentionally choose the images that are self-occluded,
truncated, or in poor lighting conditions. The results show that our method
is competent to generate appealing and robust hand meshes.
\label{sec:result}

\vspace{-5 pt}
\subsection{Ablation Study}
\label{sec:ablation}
We conduct this ablation study to better understand the impact of different components in our network.
Specifically, we evaluate:
0) our full architecture ( {\bf ours: full model} ) in comparison to another three settings:
1) without SIKNet ( {\bf ours: w/o SIK} );
2) without SIKNet and depth predictor ( {\bf ours: w/o SIK\&D} );
3) without SIKNet, depth and silhouette predictor ({\bf ours: w/o SIK\&D\&S}).
The following comparisons are controlled by only one variable. 0) $vs$ 1) shows the accuracy improvement of adding SIKNet. 1) $vs$ 2) and 2) $vs$ 3) also show the improvement of employing depth and silhouette decoder as bisected branch, respectively.
These experiments are conducted only on RHD dataset since the STB contains little variation and is easily saturated.  As a baseline, we choose the result reported by Zhang \etal\cite{zhang2019end}. It is shown in Fig. \ref{fig:quantitative} (right column) that our method outperforms theirs by a large margin.
The result is consistent with our expectation that employing silhouette and depth map predictor can greatly help lift the pose estimation from 2D to 3D, and employing hand pose prior through SIKNet can further correct noise estimation.

\vspace{-5 pt}
\section{Conclusion}
In this paper, we proposed a novel method to address the challenging task of 3D hand mesh reconstruction from a single RGB image, named BiHand. Since the mesh is highly irregular and hard to be directly optimized, the procedures in BiHand are decoupled into 3 stages, i.e. 2D seeding, 3D lifting, and mesh generation stage.  We have argued that 2D keypoints and silhouette, 3D joints and depth map, joint rotations and shape parameters, are closely related yet different in details, thus we adopted a stacked bisected design across the whole network. Experiments have showed the efficacy of such design on both 3D hand estimation and mesh reconstruction. In the future, we will adapt BiHand to hand interacting with an object, and further exploit the bisected design in that scenario.

\bibliography{egbib}

\end{document}